\pdfoutput=1

\documentclass[11pt]{article}

\usepackage{acl}
\usepackage{times}
\usepackage{latexsym}

\usepackage[T1]{fontenc}

\usepackage[utf8]{inputenc}
\usepackage{CJKutf8}
\usepackage{microtype}

\usepackage{mathptmx} 
\usepackage{latexsym}
\usepackage{xcolor}
\usepackage{stackengine}

\usepackage{adjustbox}
\usepackage{subfigure}
\usepackage{algorithm2e}
\usepackage{setspace}
\usepackage{breqn}
\usepackage{bbm}
\usepackage{dirtytalk}
\usepackage{graphicx}
\usepackage{float}
\usepackage{listings}
\usepackage{booktabs}
\usepackage{caption}
\usepackage{enumitem}
\usepackage{amsmath,stackengine}
\usepackage{cleveref}
\usepackage{float}
\stackMath

\title{Local Structure Matters Most in Most Languages} 

\author{Louis Clouâtre\textsuperscript{1,3}
  Prasanna Parthasarathi\textsuperscript{2}
  Amal Zouaq\textsuperscript{1} and
  Sarath Chandar \textsuperscript{1,3,4} \\
  \textsuperscript{1} Polytechnique Montréal\\
  \textsuperscript{2} Noah's Ark Lab, Huawei Canada\\
  \textsuperscript{3} Quebec Artificial Intelligence Institute (Mila) \\
  \textsuperscript{4} Canada CIFAR AI Chair 
}

\begin{document}
\maketitle
\begin{abstract}
Many recent perturbation studies have found unintuitive results on what does and does not matter when performing Natural Language Understanding (NLU) tasks in English.
Coding properties, such as the order of words, can often be removed through shuffling without impacting downstream performances.
Such insight may be used to direct future research into English NLP models.
As many improvements in multilingual settings consist of wholesale adaptation of English approaches, it is important to verify whether those studies replicate or not in multilingual settings.
In this work, we replicate a study on the importance of local structure, and the relative unimportance of global structure, in a multilingual setting.
We find that the phenomenon observed on the English language broadly translates to over 120 languages, with a few caveats.

\end{abstract}

\section{Introduction}

A recent research trend has explored the sensitivity, or insensitivity, of neural language models to different perturbations of texts~\citep{pham2020out,sinha2020unnatural, sinha2021masked, gupta2021bert,o2021context,ShakingTrees, clouatre-etal-2022-local}.
Their findings may be central in directing future NLP research by providing insight into which coding property~\citep{coding_properties} of language are most valuable to performing Natural Language Understanding (NLU) tasks.
As research in English NLP tends to be adapted to other languages, such as through single language adaptation of BERT-style models~\citep{BERT, ChineseBERT1, FrenchBERT1, FrenchBERT2, ArabicBERT1, BrazilianT5, DutchBERT, SwedeBERT, ItalianBERT, VietBERT} or multilingual adaptations of the same architecture~\citep{XLM, Canine, mT5, ByT5, mBART, BERT}, it is vital that we verify how insights derived from the English language generalize to other languages.

One such coding property, the local structure of text, has recently been shown to be ubiquitously relied upon by both neural language models~\citep{clouatre-etal-2022-local} and humans~\citep{local_human} to understand text in English.
The global structure of text only sometimes being necessary for a model to perform NLU tasks~\citep{clouatre-etal-2022-local}.
Such results motivate hierarchical approaches to neural language model development, where one would first build meaning locally and then reason over the global context if necessary.
However, we must verify that the importance of that coding property is not merely an artifact of the English language.

In this short paper, our contributions are as follows:
\begin{itemize}
    \item We adapt and replicate the findings of \citet{clouatre-etal-2022-local} in a multilingual setting to verify their generality and find that their conclusions regarding both local and global structure broadly apply to most of the 120 languages surveyed.
    \item We provide analysis for why text using Chinese Characters as its script may be more resilient to local perturbations and highlight the importance of testing improvements in English neural modeling in other languages.
\end{itemize}

\section{Related Work}
\label{sec:related-work}

\paragraph{Text Perturbations and Structure Probing} 

Several text perturbation schemes have been explored to probe what kind of structure does and does not matter for neural models performing NLU.
\citet{sankar-etal-2019-neural} explores both shuffling and reversing utterances and words in a generative dialogue setting, highlighting models' insensitivity to the order of conversational history.
\citet{pham2020out} explores shuffling $n$-grams for different values of $n$, which highlights the insensitivity of pretrained Transformer models.
\citet{sinha2020unnatural} explores shuffling of words on textual entailment tasks, highlighting models' insensitivity to such perturbations.
Finally, \citet{ShakingTrees} extend perturbation studies to Swedish and Russian and performs perturbations by shuffling syntactic phrases, rotating sub-trees around the root of the syntactic tree of a sentence, or simply shuffling the words of the text.

These approaches share the main limitation of requiring automatic parsing tools or well-developed tokenizers to define words.
This limits their applicability in a multilingual setting.
Priors regarding the form of the text, such as the presence of white-space delimited words, limit the generalizability of most of these studies.

\citet{clouatre-etal-2022-local} proposes a suite of controllable perturbations on characters and subwords, which should be compatible with almost any written language, as well as a metric quantifying perturbations to the local and global structure that measures perturbations on a character-level.

\section{Experiments}
We extend the perturbation studies of \citet{clouatre-etal-2022-local} to a multilingual setting.
We perform those experiments on eight popular cross-lingual tasks~\citep{XTREME, XCOPA, XGLUE} covering over 120 languages.
This will shed light on what languages, if any, do not share the same sensitivity to local structure and insensitivity to global structure as English.

\subsection{Metric and Perturbations}

The \textbf{CHRF-2} (chrF)~\citep{chrf} metric measures the amount of character bi-gram overlap between a perturbed text and the original text.
This measure represents the amount of \textit{local} structure that has not been perturbed in a text.

The \textbf{I}ndex \textbf{D}isplacement \textbf{C}ount (IDC)~\citep{clouatre-etal-2022-local} metric measures the average absolute distance traversed by every character in a perturbed text.
An IDC of $0.3$ would mean that, on average, every character has traversed $30\%$ of the length of the text.
This measure represents the amount of \textit{global} perturbations applied to a text.

The \textbf{compression rate} (Comp)~\citep{ByT5} represents the total length of the text in terms of characters divided by the total length of the text once tokenized.
Since most of our models either use subwords or tokenize characters directly, there are no out-of-vocabulary tokens to be counted.
The compression rate is then used as a proxy for vocabulary destruction of pretrained models, an important confounder for the importance of local structure.

\begin{figure}[ht]
    \centering
    \includegraphics[width=0.95\columnwidth]{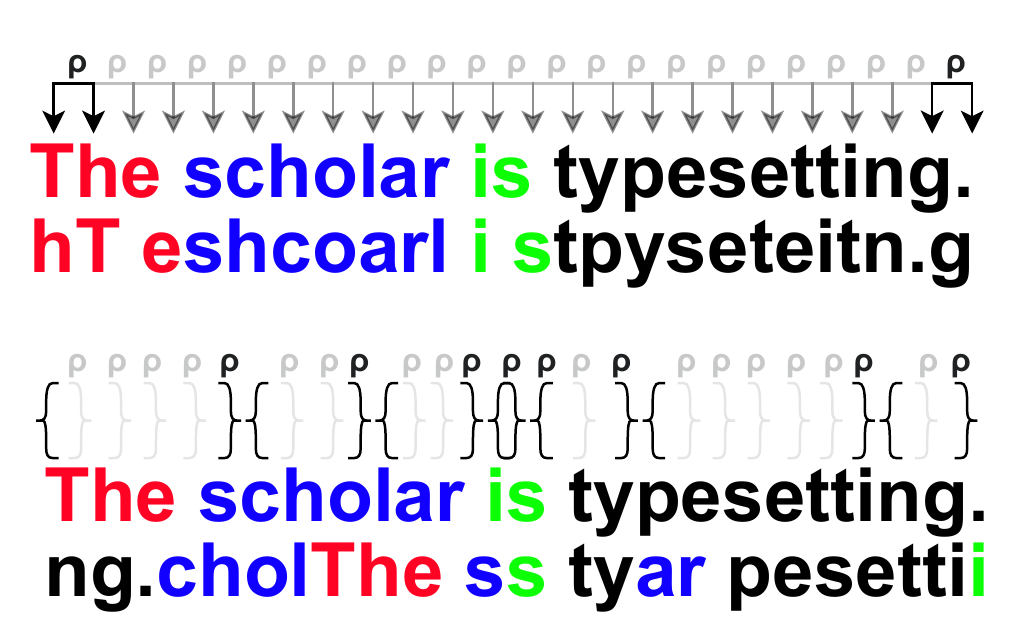}
    \caption{From top to bottom: Neighbor Flipping with $\rho=0.5$, Phrase Shuffling with $\rho=0.5$}
    \label{fig:sample-perturbations}
\end{figure}

            
            
            
           

We perform perturbations by altering the order of \textbf{subwords} and \textbf{characters} present in the text. 
Three types of perturbations are applied.

\textbf{Full shuffling} completely randomizes the order of the subword or characters.

\textbf{Neighbor flipping} flips a subword or character with its neighbor with a controllable probability $\rho$, providing local perturbations while maintaining much of the absolute position of the tokens.

\textbf{Phrase shuffling} randomly builds phrases of subwords or characters of controllable average length with a parameter $\rho$ and shuffles those phrases, providing a minimal amount of local perturbations for a large amount of change in absolute position.

Simple examples of those perturbations are shown in Figure~\ref{fig:sample-perturbations}, pseudocode and details are present in the Appendix~\ref{app:pseudocode_perturbations}.


\begin{table}[ht]
\centering
\begin{adjustbox}{width=1\columnwidth}
\small
\begin{tabular}{||c c c c||} 
 \hline
 \textbf{Task} & $n$ Languages & Task Type & Metric\\ [0.5ex] 
 \hline\hline
 \textbf{PAWS-X} & 7 & Paraphrase Detection & ACC\\ 
 \hline
 \textbf{XNLI} & 15 & NLI & ACC\\ 
 \hline
 \textbf{QAM} & 3 & Text Classification & ACC\\ 
 \hline
 \textbf{QADSM} & 3 & Text Classification & ACC\\ 
 \hline
 \textbf{WPR} & 7 & Page Ranking & nDCG\\ 
 \hline
 \textbf{XCopa} & 11 & Commonsense Reasoning & ACC\\ 
 \hline
 \textbf{BUCC} & 5 & Sentence Retrieval & F1\\ 
 \hline
 \textbf{Tatoeba} & 122 & Sentence Retrieval & F1\\ 
 \hline

\end{tabular}

\end{adjustbox}
\caption{Summary information of the different tasks used.} 
\label{tab:task_statistics}
\end{table}

\begin{figure*}[h!t]
    \centering
    
    \includegraphics[width=0.35\textwidth]{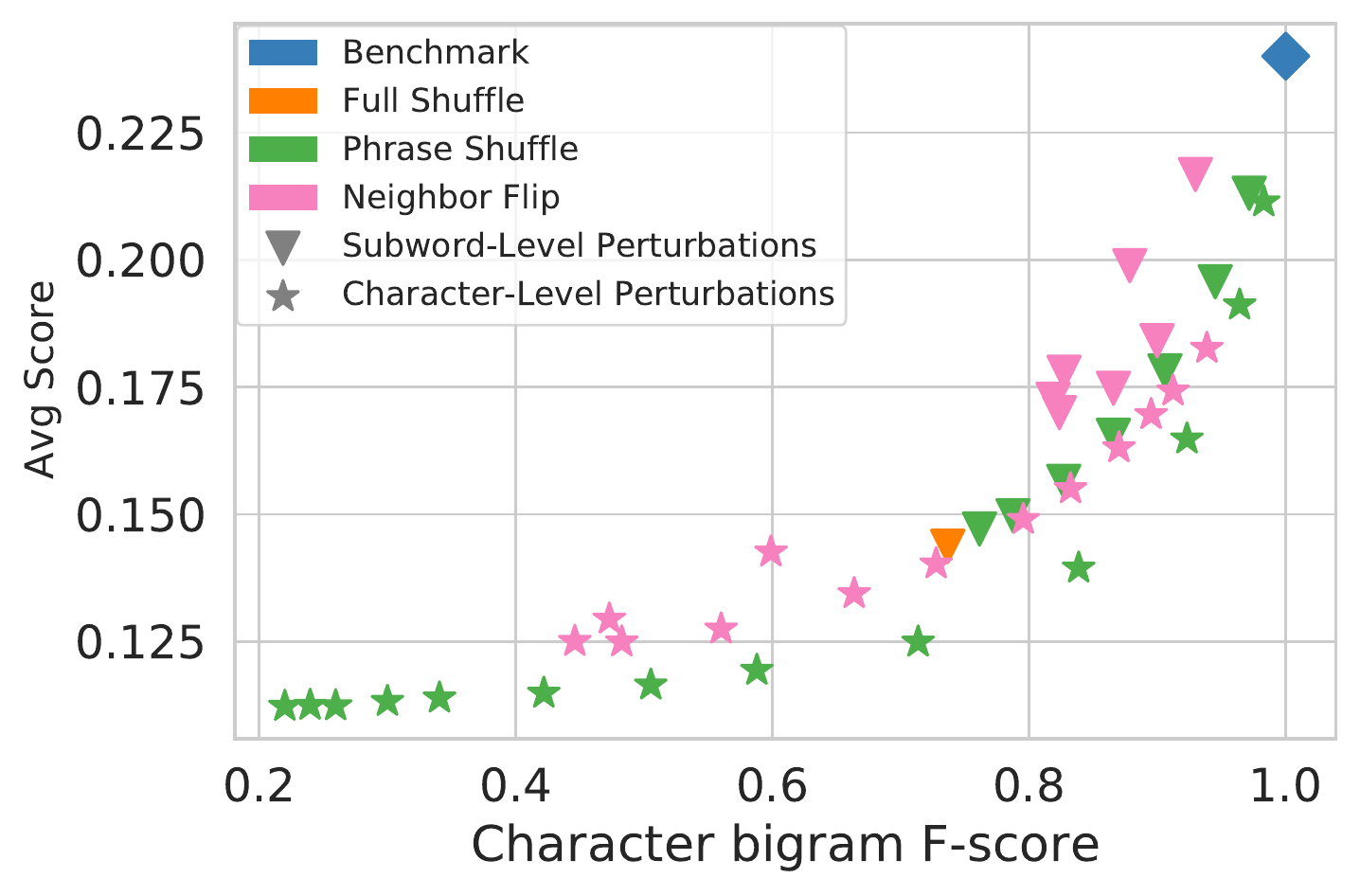}\hfill
    \includegraphics[width=0.31\textwidth]{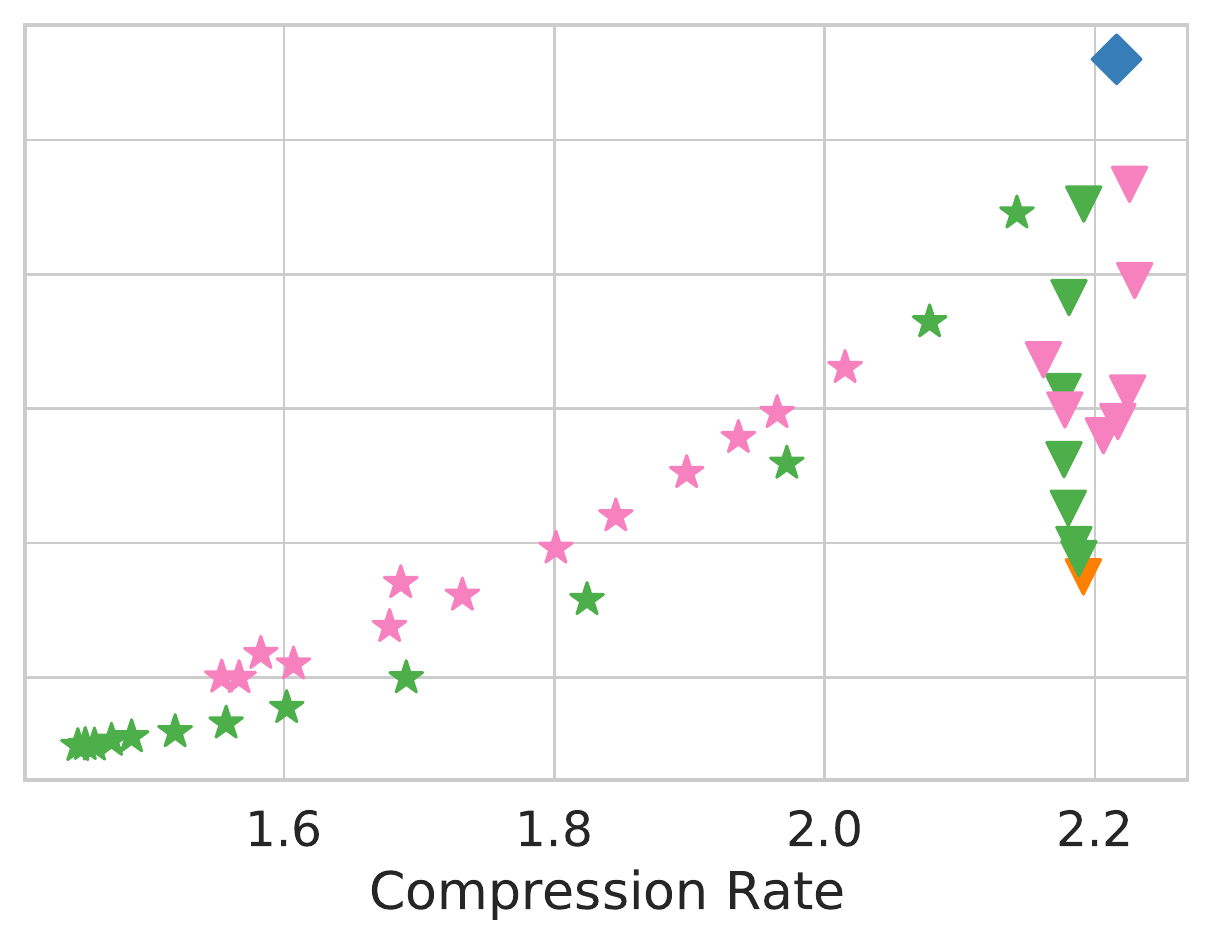}\hfill
    \includegraphics[width=0.31\textwidth]{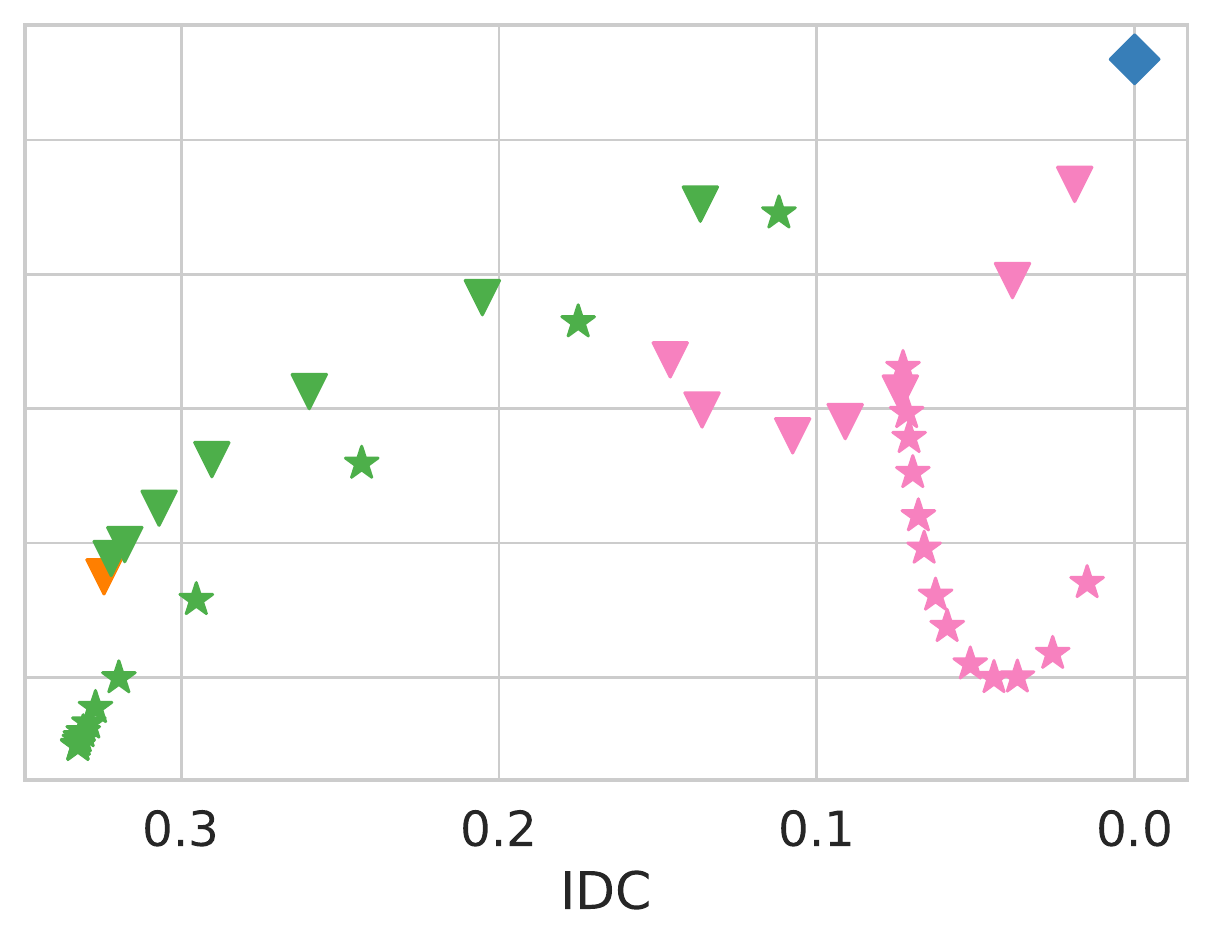}

    \caption{
    Plotted are the relations between the different choices of metrics measuring the amount of perturbation and the average performance of all 3 models on all tested datasets.
    Left is more perturbed, up is better performance.
    The X-axis of the IDC metric is inverted for clearer comparison.
    }
    \label{fig:global_metrics}
\end{figure*}

\subsection{Experimental Details}
All experiments are conducted on three pretrained cross-lingual models.
The XLM-RoBERTa-Base~\citep{XLM}, BERT-Base-Multilingual-Cased~\citep{BERT} and the Canine-S~\citep{Canine} model are used.
The Canine model is a tokenization-free pretrained model, which lets us isolate the impact of subword destruction on the findings.

The zero-shot cross-lingual setting~\citep{XTREME} is used for all experiments.
The model is first finetuned on the English version of the dataset and evaluated without further tuning on all target languages.

The English version on which the model is finetuned is kept unperturbed, while the target language text on which the model is evaluated goes through several perturbations.
We perform a total of 43 different perturbations on every task and language and obtain their performance.
All models are finetuned on five different random seeds, and all perturbations are performed on five different random seeds, for a total of 25 evaluations for every model on every task, every language present in the tasks, and every perturbation setting.~\footnote{Detailed training and testing hyperparameters and process are present in the Appendix \ref{app:hyperparameters_training} and details on the specific perturbations in Appendix~\ref{app:perturbations}.}

A total of 8 cross-lingual tasks selected from the most popular cross-lingual benchmarks~\citep{XTREME, XGLUE, XCOPA} covering over 120 languages are used for evaluation.~\footnote{Extractive tasks such as extractive QA are not compatible with our perturbations, as the answer would also be perturbed and were not considered.}
Summary information of the tasks can be found in Table~\ref{tab:task_statistics}.~\footnote{As we use all 122 languages in the Tatoeba dataset, which vary from 100 to 1000 possible sentences to retrieve, the F1 score is more appropriate as an evaluation of performance than the accuracy used in the XTREME benchmark.}

\subsection{Results and Discussion}

\begin{figure}[ht]
    \centering
    \includegraphics[width=0.95\columnwidth]{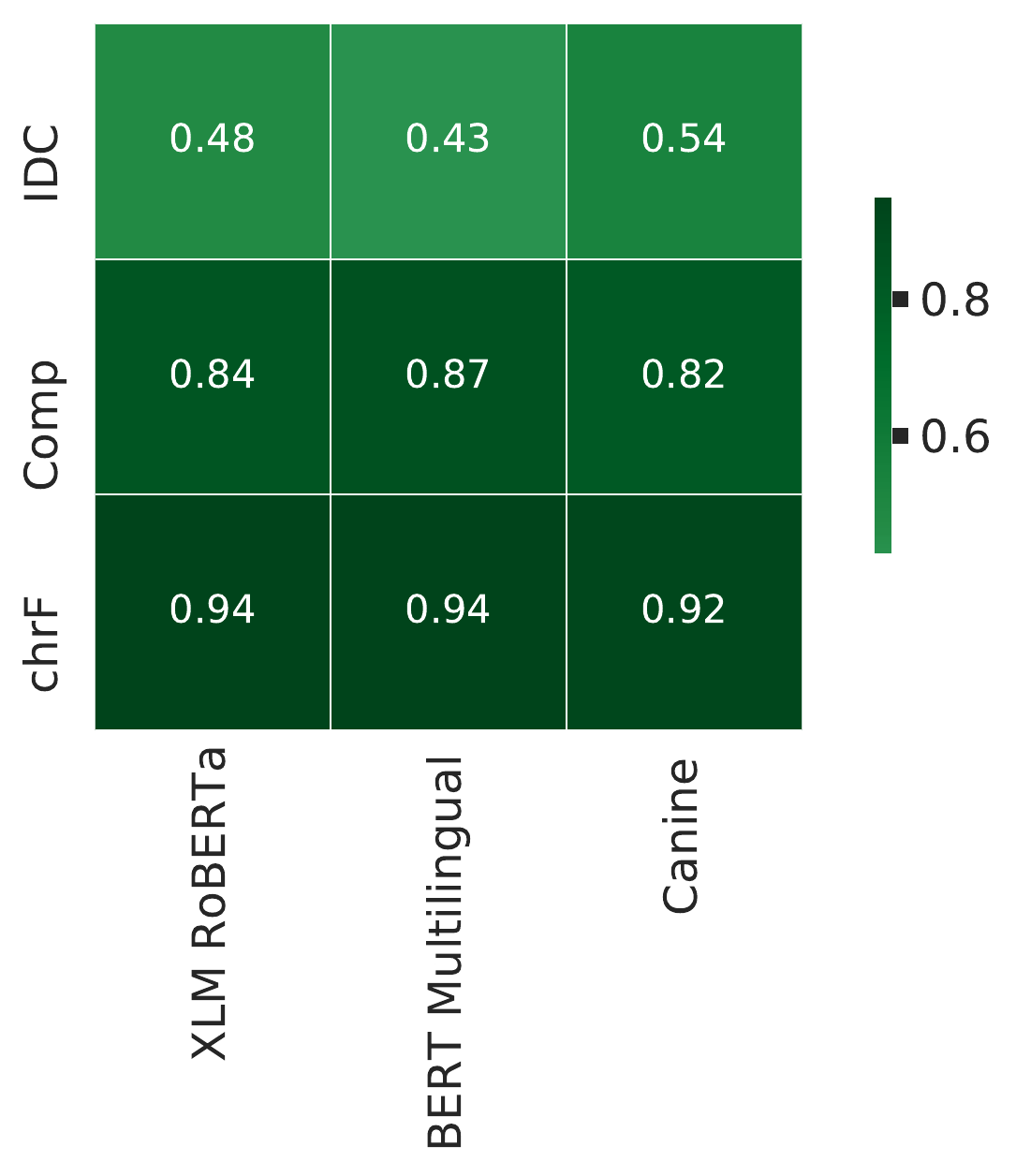}
    \caption{Rank-correlation matrix between the different models' performance to perturbed samples on the and the perturbation quantified by the different metrics.
    The higher the value the better the metric explains the degradation in performance.}
    \label{fig:models-metric-correlation}
\end{figure}

\begin{figure}[ht]
    \centering
    \includegraphics[width=0.95\columnwidth]{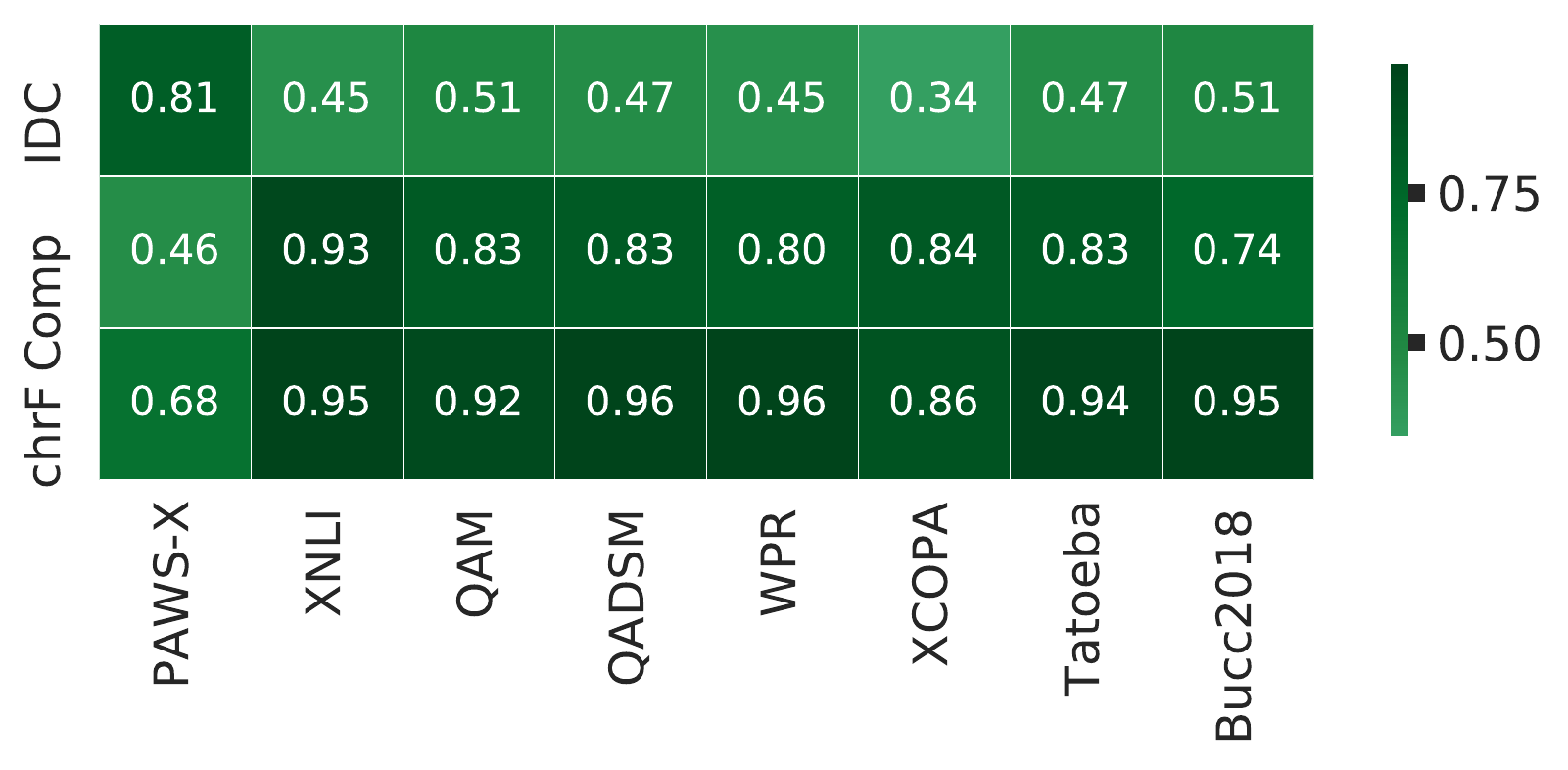}
    \caption{Rank-correlation matrix between the different task's performance to perturbed samples and the perturbation quantified by the different metrics.
    The higher the value the better the metric explains the degradation in performance.}
    \label{fig:task-metric-correlation}
\end{figure}

In Figure~\ref{fig:global_metrics}, we observe the trends reported by \citet{clouatre-etal-2022-local} to be broadly true in a cross-lingual setting.
Specifically, the more local perturbations are applied to a text, the more degradation in the understanding of that text can be expected, which shows that model does rely on the local structure to build understanding.
The perturbations to the global structure are shown to be a much poorer explanation for the degradation in performance than the perturbation to the local structure.
The compression rate is highly correlated with a model's performance and the local structure, making it a potential confounder for the degradation in performance.
However, the trend in local structure holds with subword-level perturbations, unlike with the compression rate, which is not affected by perturbations to the order of subwords, as well as holding for the vocabulary-free Canine model, as shown in Figure~\ref{fig:models-metric-correlation}.
This makes it more likely that the cause for the degradation in performance is the local structure perturbation, the destruction of the vocabulary being incidental.

\subsubsection{PAWS-X}
Figure~\ref{fig:task-metric-correlation} shows the rank-correlations of a model's performance over the different tasks with the different measures of perturbation.
The overall trends are stable in all but one task, PAWS-X.
Much like the CoLA task~\cite{glue1} in the GLUE Benchmark~\cite{GLUE}, it is possible to build tasks that require the specific order of words to be successfully completed.
The PAWS-X task comprises adversarial paraphrases containing a similar lexicon between paraphrase and non-paraphrases.
The performance is then highly sensitive to perturbations causing displacement, such as shuffling words, even if the local structure is mostly kept intact.
It is not that local structure is unnecessary, but that global structure is.
This phenomenon is further explored by \citet{cola_paws_1, cola_paws_2, cola_paws_3}.



\subsubsection{Chinese Character Script}\label{sec:chinese_characters}
\begin{figure}[h!t]
    \centering
    \includegraphics[width=0.95\columnwidth]{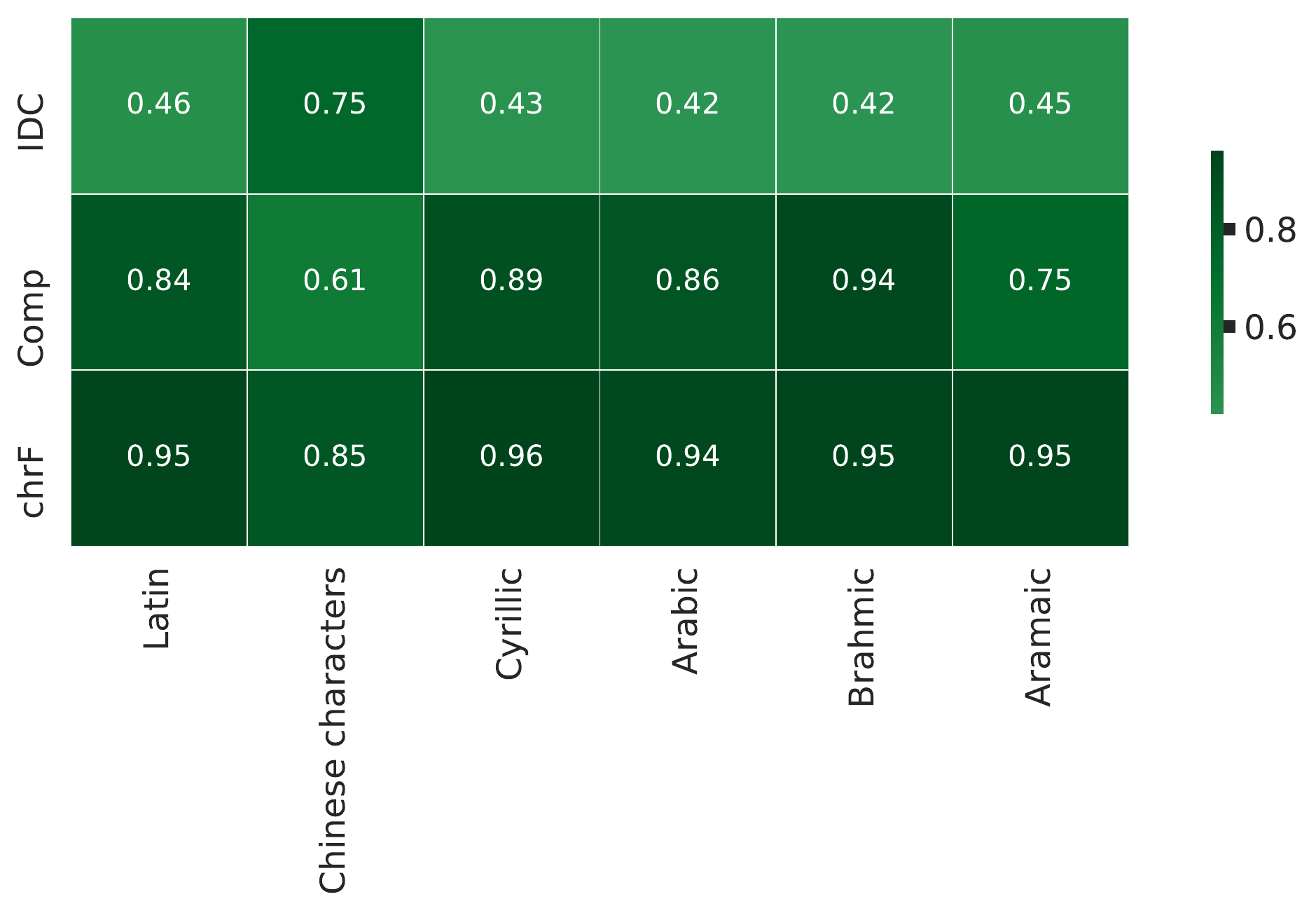}
    \caption{Rank-correlation matrix between the different language script's containing at least 3 languages performance to perturbed samples on the and the perturbation quantified by the different metrics.
    The higher the value the better the metric explains the degradation in performance.}
    \label{fig:script-metric-correlation}
\end{figure}

Figure~\ref{fig:script-metric-correlation} show that the findings are consistent across almost all text scripts, with the exception of languages using Chinese Characters as script.


This is most likely caused by how semantically richer the smallest separable unit in Chinese tends to be compared to characters in different scripts.
Where Chinese has a single indivisible character meaning "water" the English equivalent "water" can be perturbed to "rtawe".
Even character-level shuffling cannot strip Chinese text of all meaning, which would explain some the differences.
It is to be noted that while weaker, the correlation between local structure perturbations and performance remains high.

\section{Conclusion}

We first explored and confirmed the importance of local structure, the limited importance of global structure, and controlled for the potential of vocabulary destruction being the main explanatory factor in 8 NLU tasks covering over 120 languages.
In aggregate, the findings of \citet{clouatre-etal-2022-local} hold for many different pretrained cross-lingual models and NLU tasks in a multilingual setting.
Local structure sensitivity and global structure insensitivity do not seem to be an artifacts of the English language.

A significant exception is when grammatical cues are essential to complete the task, such as in the PAWS-X task.
While many tasks can be solved purely with the information obtained from the local structure, reasoning over the global context is necessary for many problems.

Languages using Chinese characters as their script also deviate from the norm.
This is likely caused by how semantically rich their characters are.

It will be important that any NLP improvements derived from English experiments are verified to also generalize to other languages.
As we have observed that languages written in Chinese Character Script are differently impacted by perturbations to different coding properties, it is possible that improvements to the way our model understand those properties in English will not generalize.

\section*{Acknowledgements}
This research has been funded by the NSERC Discovery Grant Program.

\newpage
\bibliography{anthology,custom}
\bibliographystyle{acl_natbib}

\appendix

\onecolumn
\section{Experiment Details}\label{app:experiments}

\paragraph{Model Hyperparameters and Training}\label{app:hyperparameters_training}

We finetune each pretrained models on the English version of each dataset for a total of 10 epochs, checkpointing the model after each epochs.
The English version is never perturbed, the finetuning is done on unperturbed data.
This finetuning is done 5 times with different random seeds for each model and each datasets.
For 8 datasets and 3 models we have a total of $3 * 8 * 5 = 120$ finetuning and $1200$ checkpoints, one for each epoch.
A learning rate of 2e-5, a batch size of 32 and a weight decay of 0.1 is used in all finetuning.
All experiments used a warmup ratio of 0.06, as described in \citet{RoBERTa}.

For the evaluation, we perform the same perturbations on the validation and testing data of the different target languages.
We evaluate the perturbed validation data on each of the 10 checkpoints, chose the best checkpoint on the perturbed validation data, and evaluate that checkpoint on the perturbed test data.
This process is repeated for each perturbations, each of the 5 random seed and 5 times with different perturbation random seeds for each finetuned models.
In total, for each language in each task on each model for each perturbation setup we average results over 25 random seeds.

For the sentence retrieval tasks, such as Tatoeba, we do not perform any finetuning.
We simply obtain the nearest neighbour using cosine similarity on the final hidden representation.~\citep{XTREME}
First, we obtain the representation of the unperturbed English side of the dataset.
This is done by feeding the English text through the model and averaging the final layers hidden representation of the text.
We then perform our perturbations on the target language text, feed those perturbed text through the same pretrained cross-lingual model and obtain it's representation through the same process.
We now have a set of English representation and a set of target language representation, on which we find the nearest neighbour as measured by the Cosine Distance on the pooled hidden representations.
If the nearest neighbour is the sentence that was to be retrieved, we consider this an hit, else it is a miss.
The reported results are over the average of 5 random seeds of those perturbations.

\paragraph{Perturbations}\label{app:perturbations}
A total of 43 perturbations are used for all experiments.
The first one is the Benchmark, which is simply the unperturbed text.
We perform a full-shuffling on both the subwords and characters.
On the subword-level perturbations we perform phrase-shuffling with $\rho$ values of: $[0.9$, $0.8$, $0.65$, $0.5$, $0.35$, $0.2$, $0.1]$ and neighbour-flip shuffling with $\rho$ values of: $[0.9$, $0.8$, $0.6$, $0.5$, $0.4$, $0.2$, $0.1]$.
On the character-level perturbations we perform phrase-shuffling with $\rho$ values of: $[0.975$, $0.95$, $0.9$, $0.8$, $0.65$, $0.5$, $0.4$, $0.3$, $0.2$, $0.15$, $0.1$, $0.075$, $0.05]$ and neighbour-flip shuffling with $\rho$ values of: $[0.8$, $0.65$, $0.5$, $0.4$, $0.3$, $0.2$, $0.1$, $0.075$, $0.05$, $0.035$, $0.025$, $0.01]$.
A total of 15 subword-level experiments, 27 character-level experiments and the unperturbed benchmark are evaluated for a grand total of 43 different perturbation settings .

\newpage
\section{Pseudocode for Metric and Perturbations}
\label{app:pseudocode_perturbations}
\newcommand{\myalgorithma}{%
      \begin{algorithm}[h]
      \small
      \SetAlgoLined
      \SetKw{Kw}{\KwInput}
      \SetKwInOut{Input}{input}
      \SetKwInOut{Output}{output}
      \SetKwProg{Fn}{Function}{:}{\KwRet}
      
      \SetKwFunction{Func}{IDC}
      \Fn{\Func{$X_p$}}{
       $X_p^{len} \leftarrow X_p.\texttt{length()}$\;
      IDC\_list $\leftarrow$ \texttt{list()}
      
      \For{ $i \leftarrow\ 0$ {\bf and} $i \leq X_p^{len}$}{
      abs\_distortion $\leftarrow$ \texttt{abs(i-$X_p\left[i\right]$)}\;
      IDC\_list.\texttt{append}(abs\_distortion)\;
      }
      IDC\_agg $\leftarrow$ IDC\_list.\texttt{mean}()\;
      
      IDC $\leftarrow \frac{IDC\_agg}{X_p^{len}}$\;
        }
    \caption{Pseudocode to compute IDC metric.}
    \end{algorithm}
}
\newcommand{\myalgorithmb}{
\begin{algorithm}[h]
\small
\SetAlgoLined
  \SetKw{Kw}{\KwInput}
  \SetKwInOut{Input}{input}
  \SetKwInOut{Output}{output}
  
  \SetKwProg{Fn}{Function}{:}{\KwRet}
  
  \SetKwFunction{Func}{DND}
  \Fn{\Func{$X_p$}}{
  $X_p^{len} \leftarrow$ $X_p.\texttt{length()}$\;
  DND\_list $\leftarrow$ \texttt{list()}
  
  \For{ $i \leftarrow\ 0$ {\bf and} $i \leq X_p^{len} - 1$}{
  \eIf{$X_p[i] - X_p[i+1]==$1}{
  DND\_list.\texttt{append}($0$)\;
  }{
  DND\_list.\texttt{append}($1$)\;
  }
  }
  DND\_agg $\leftarrow$ DND\_list.\texttt{sum}()\;
  
  DND $\leftarrow \frac{DND\_agg}{(X_p^{len} - 1)}$\;
    }
\caption{Pseudocode to compute DND metric.}
\end{algorithm}
}

\newcommand{\myalgorithmc}{
\begin{algorithm}[h]
  \small
  \SetKw{Kw}{\KwInput}
  \SetKwProg{Fn}{Function}{:}{\KwRet perturbed\_text}
  \SetKwFunction{Func}{PhrasePerturbation}
  \Fn{\Func{$\rho \leftarrow 0.5$, text$\leftarrow$\texttt{list}}}{

  all\_phrases $\leftarrow$ \texttt{list()}\;
  phrase $\leftarrow$ \texttt{list(text[0])}
  
  \For{token {\bf in} $\rm{text}[1:]$}{
  p $\sim Unif\left(\left[0,1\right]\right)$\;
  \eIf{$p < \rho$}{
  all\_phrases.\texttt{append}(phrase)\;
  phrase $\leftarrow$ \texttt{list}(token)
  }{
  phrase $\leftarrow \left[
  \rm{phrase,token} \right]$\;
  }
  }
  all\_phrases.append(phrase)\;
  perturbed\_text $\leftarrow$ `'.\texttt{join}(shuffle(all\_phrases))
  }
  \caption{Pseudocode for PhraseShuffle.}
\end{algorithm}
}

\newcommand{\myalgorithmd}{
\begin{algorithm}[h]
\small
  \SetKw{Kw}{\KwInput}
  \SetKwProg{Fn}{Function}{:}{\KwRet perturbed\_text}
  \SetKwFunction{Func}{NeighborFlip}
  \Fn{\Func{$\rho \leftarrow 0.5$,text$\leftarrow$\texttt{list}}}{

  perturbed\_tokens $\leftarrow$ \texttt{list()}\;
  held\_token $\leftarrow$ \texttt{list(text[0])}
  
  \For{token {\bf in} $\rm{text}[1:]$}{
  p $\sim Unif\left(\left[0,1\right]\right)$\;
  \eIf{$p < \rho$}{
  perturbed\_tokens.\texttt{append}(held\_token)\;
  held\_token $\leftarrow$ \texttt{list}(token)
  }{
  perturbed\_tokens $\leftarrow \left[
  \rm{perturbed\_tokens,token} \right]$\;
  }
  }
  perturbed\_tokens.append(held\_token)\;
  perturbed\_text $\leftarrow$ `'.\texttt{join}(perturbed\_tokens)
  }
  \caption{Pseudocode for NeighborFlip.}
\end{algorithm}
}

\myalgorithma


\myalgorithmc

\myalgorithmd


\newpage
\section{Additional Results}\label{app:add_results}

\paragraph{Language Family}

\begin{figure*}[h!t]
    \centering
    \includegraphics[width=0.95\textwidth]{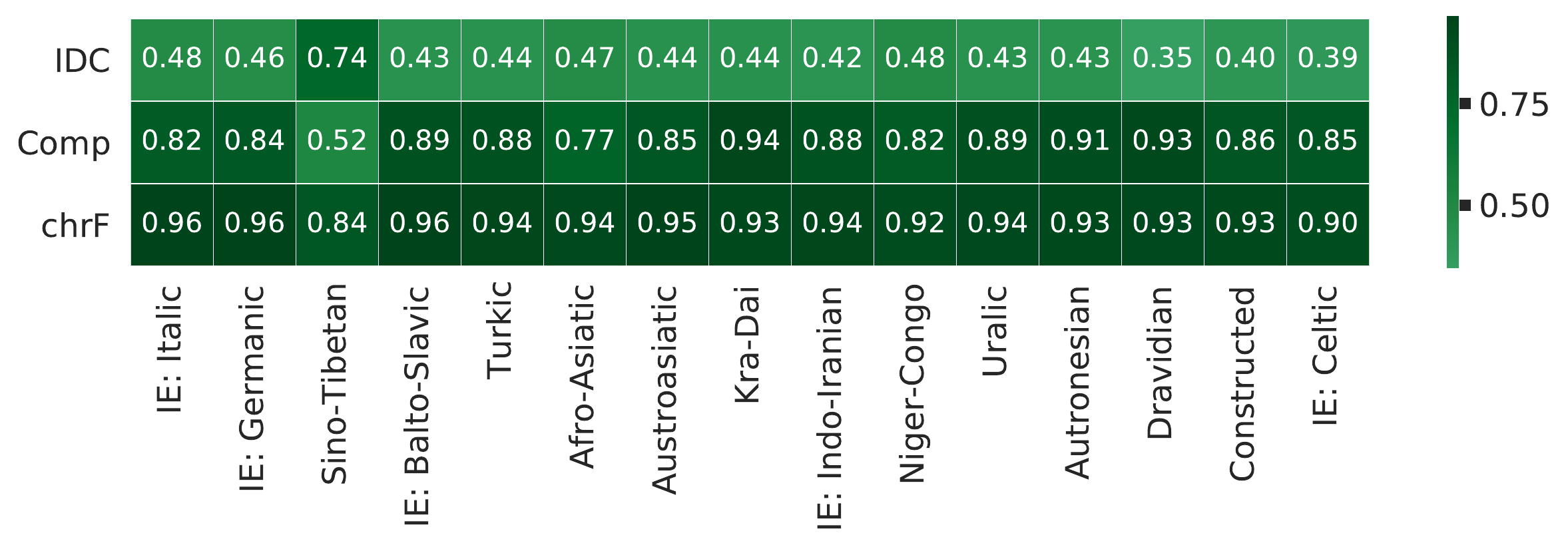}
    \caption{Rank-correlation matrix between the different language family's containing at least 3 languages performance to perturbed samples on the and the perturbation quantified by the different metrics.
    The higher the value the better the metric explains the degradation in performance.}
    \label{fig:family-metric-correlation}
\end{figure*}

Figure~\ref{fig:family-metric-correlation} shows the aggregated correlations between the different language families and the different metrics.
Results seem to be consistent across all families, with the exception of Sino-Tibetan languages.
This was generally adressed in Section~\ref{sec:chinese_characters}.

\paragraph{PAWS-X}

\begin{figure}[h!t]
    \centering
    \includegraphics[width=0.95\columnwidth]{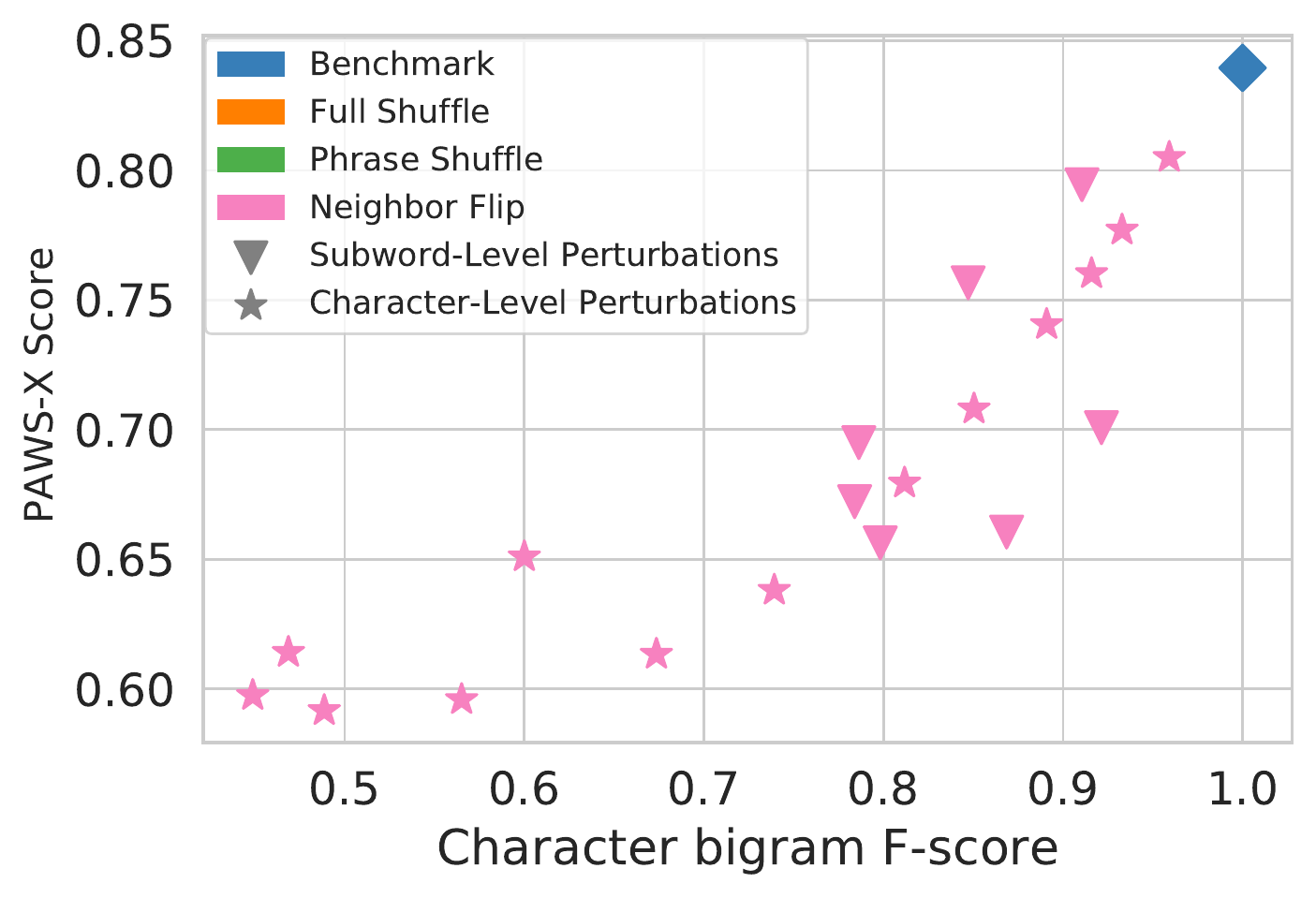}
    \caption{
    Plotted is the relations between the local structure perturbation and the average performance on the PAWS-X dataset. 
    Only the neighbour flipped perturbations are shown to isolate the impact of perturbations to the local structure.
    }
    \label{fig:paws_x_no_phrase}
\end{figure}

To determine whether it is that the local structure is not essential on PAWS-X, or simply that perturbations to the order of words are equally important, we observe the performance of models using only neighbor flipping perturbations, limiting the displacement of words to a minimum.
In Figure~\ref{fig:paws_x_no_phrase}, we show that if we only perturb the local structure, performance is highly correlated with the amount of local perturbations.
This implies that it is not that the model is insensitive to local perturbations, rather for certain tasks where grammatical queues are necessary any change to the order of words will lead to failure.  

\paragraph{Chinese Character Script}

\begin{figure*}[h!t]
    \centering
    \includegraphics[width=0.35\textwidth]{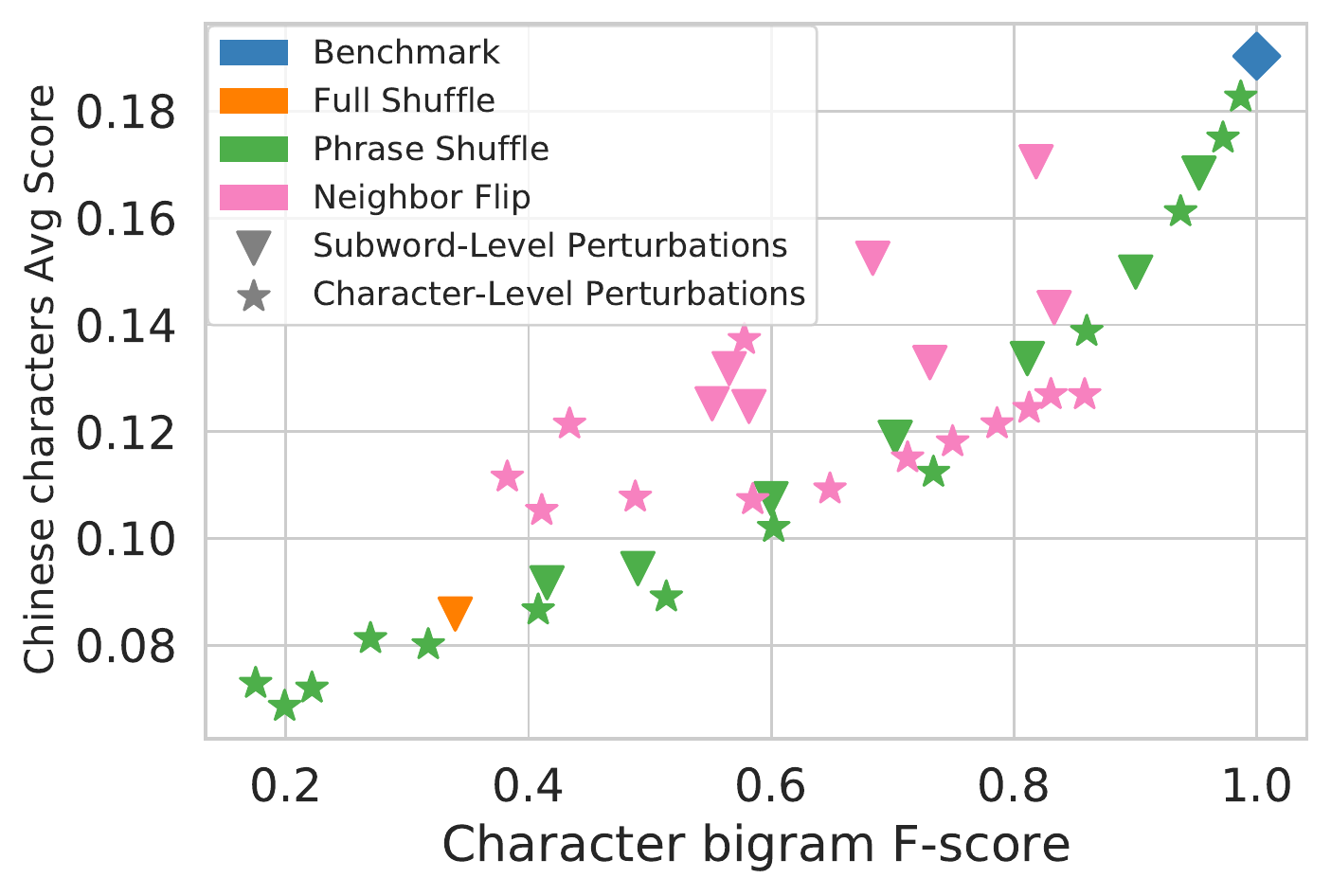}\hfill
    \includegraphics[width=0.31\textwidth]{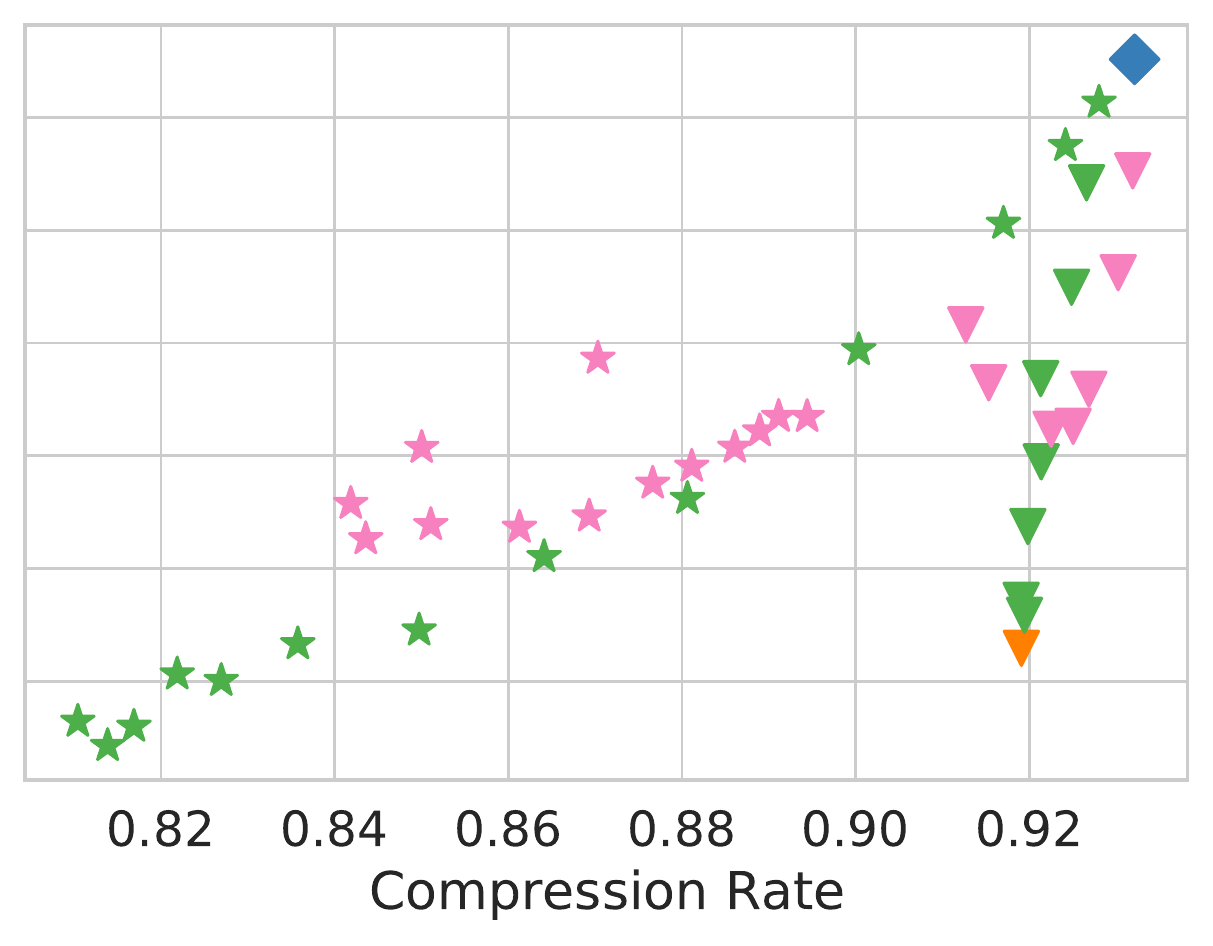}\hfill
    \includegraphics[width=0.31\textwidth]{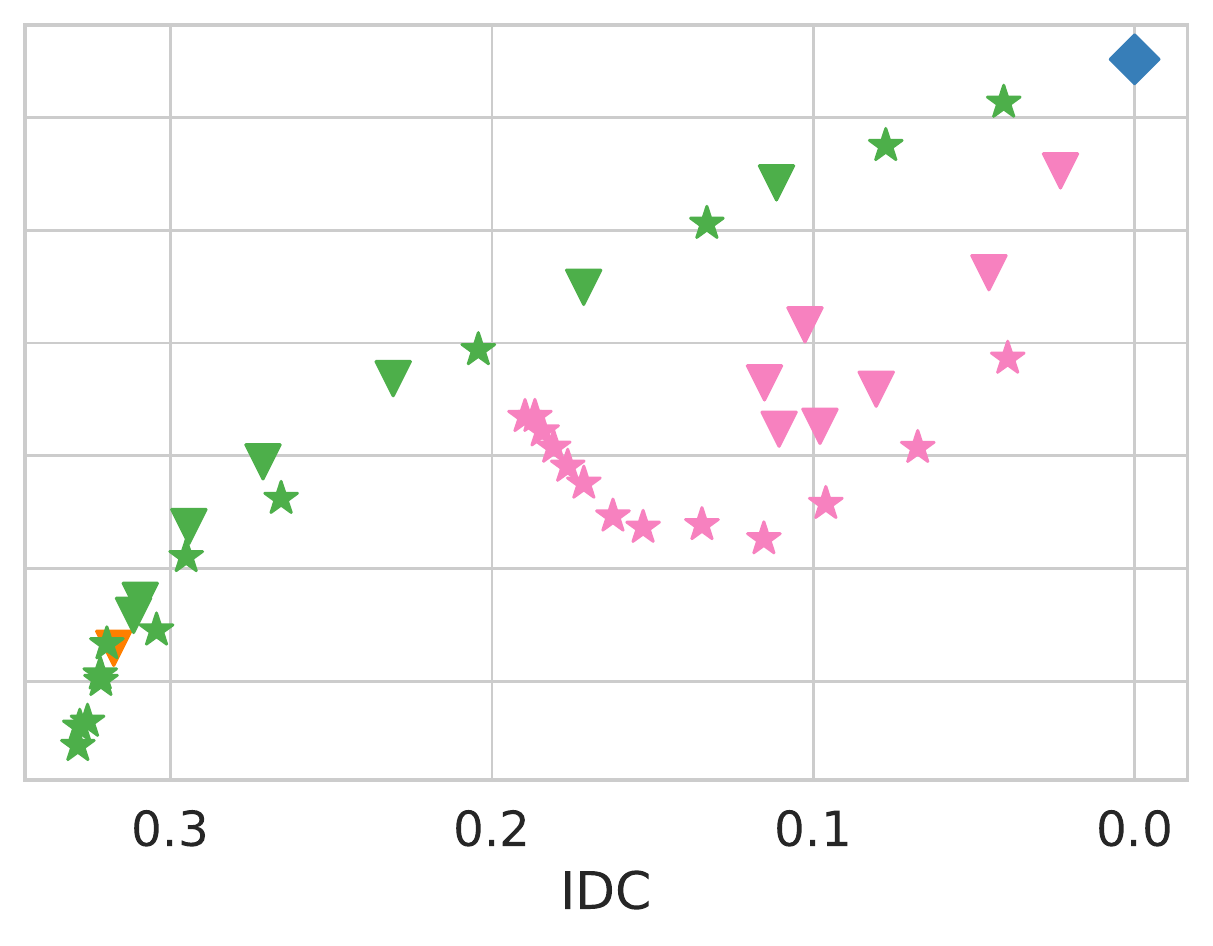}

    \caption{
    Plotted are the relations between the different metrics measuring the amount of perturbation and the average performance of all 3 models on all tested datasets on languages using chinese characters or derivatives as their scripts.
    }
    \label{fig:chinese_metrics}
\end{figure*}

\begin{figure*}[h!t]
    \centering
    \includegraphics[width=0.35\textwidth]{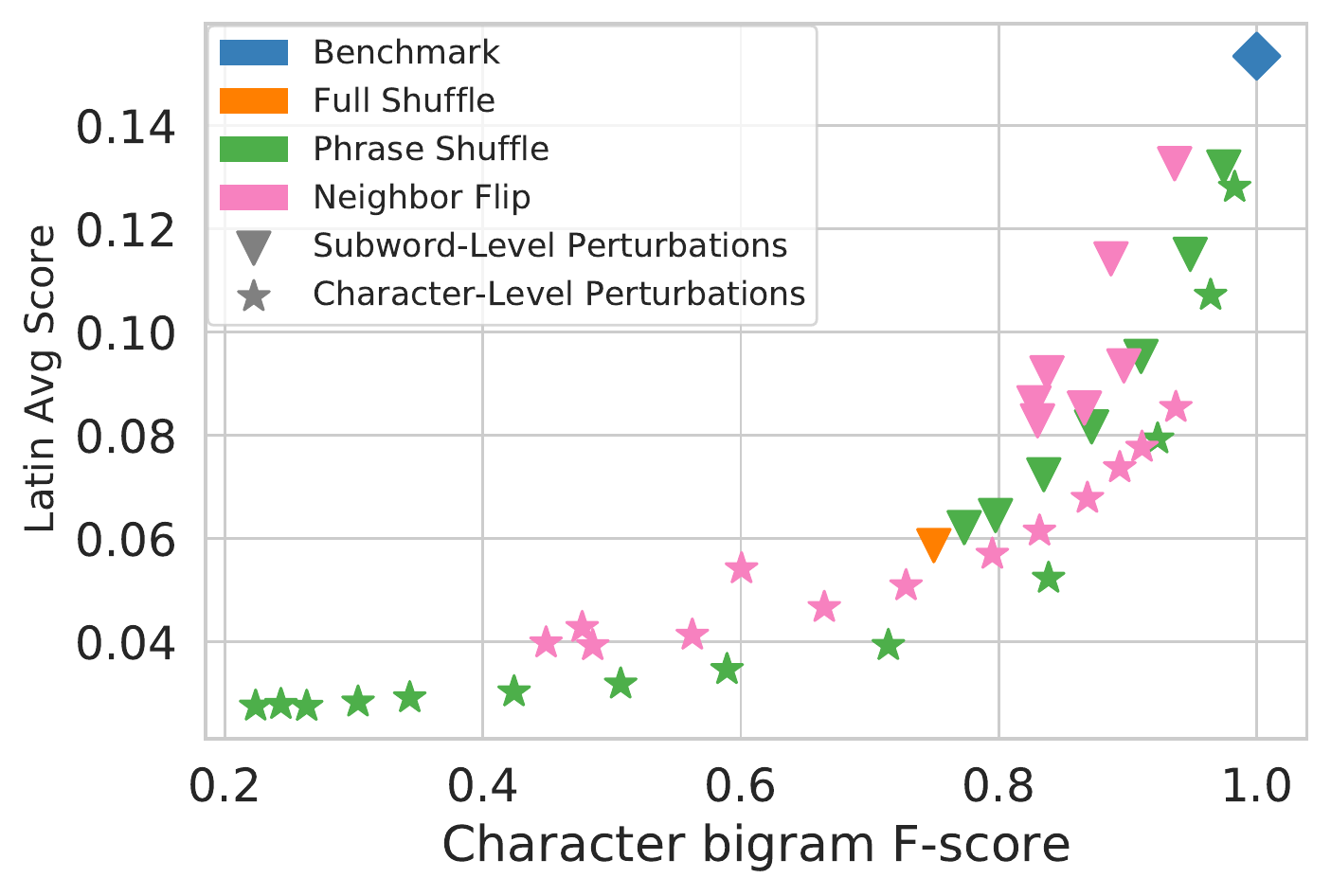}\hfill
    \includegraphics[width=0.31\textwidth]{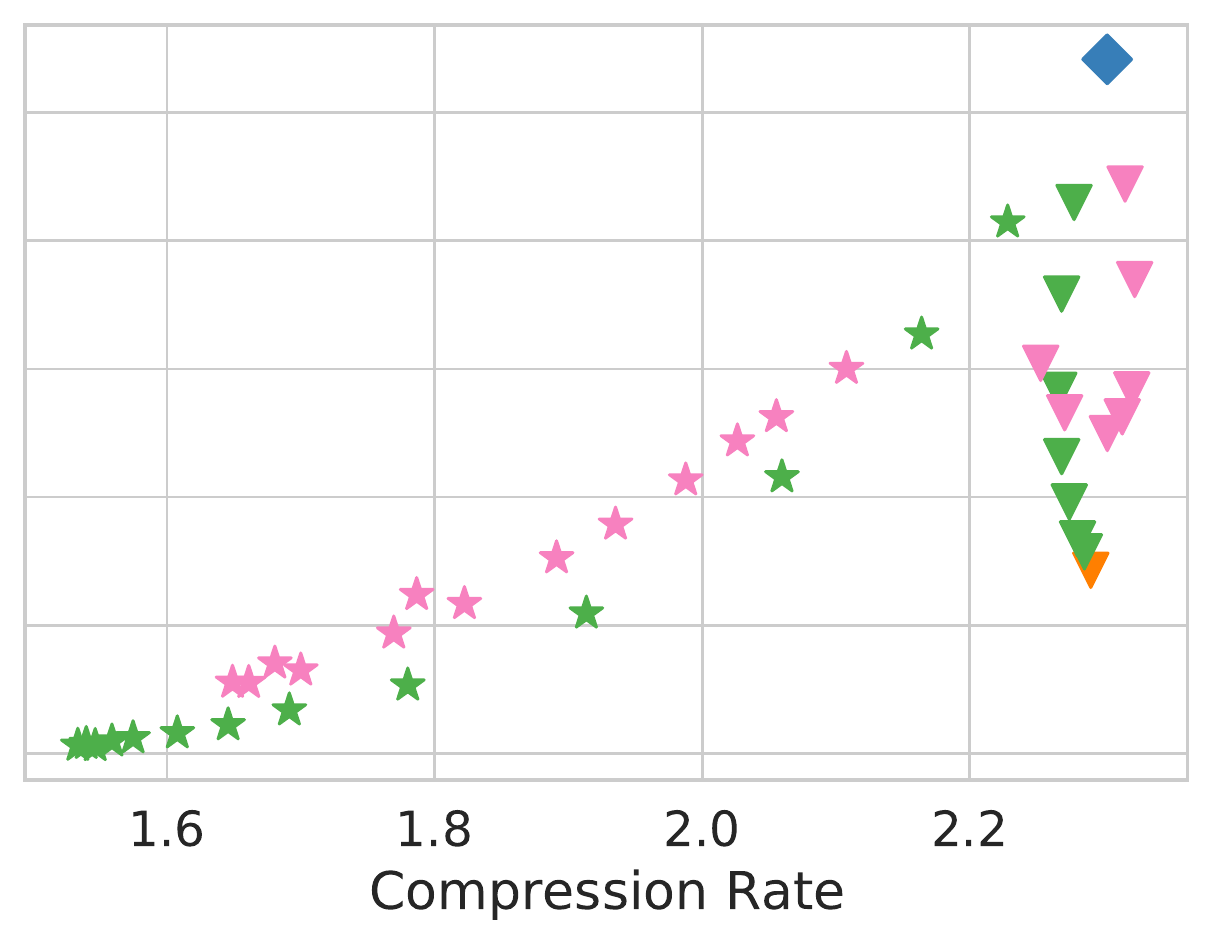}\hfill
    \includegraphics[width=0.31\textwidth]{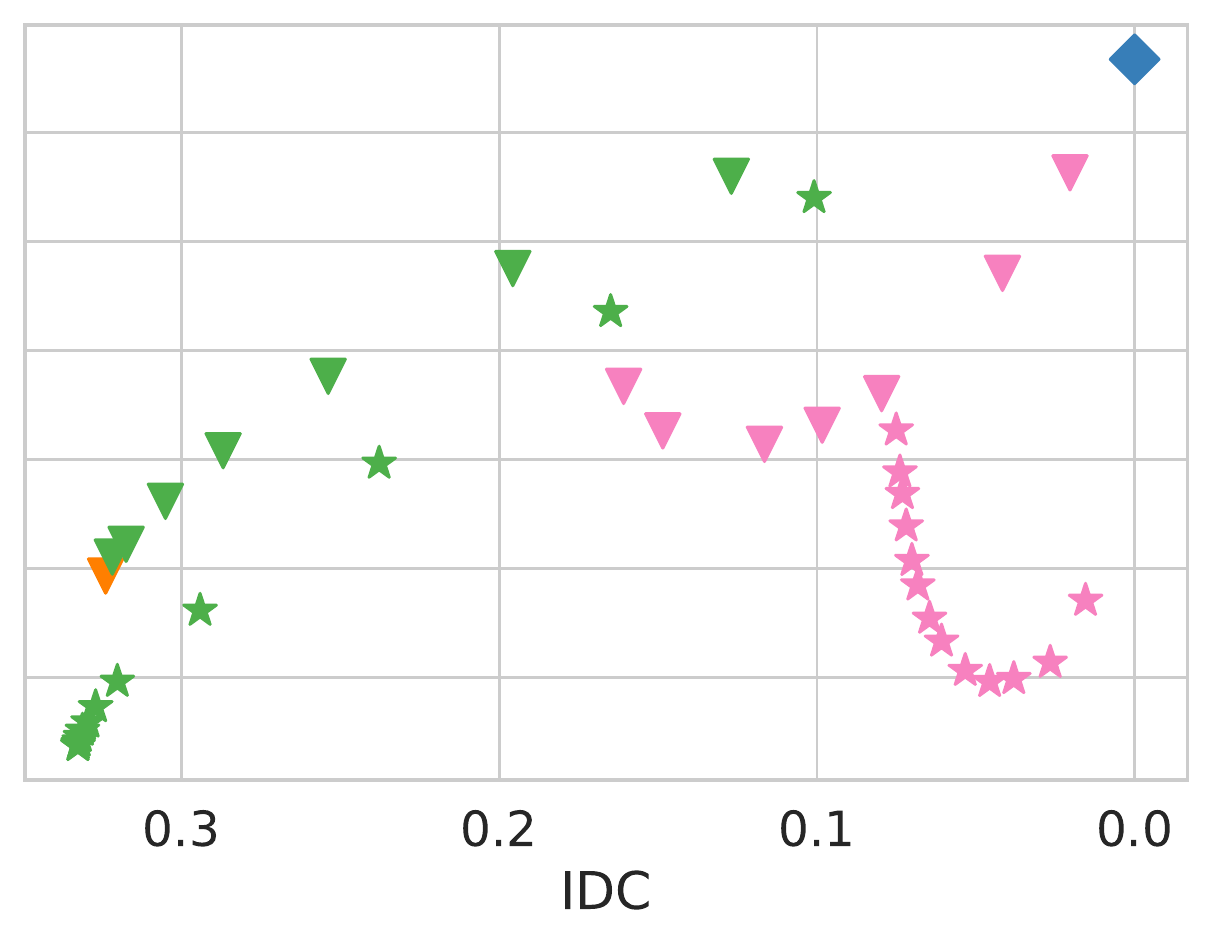}

    \caption{
    Plotted are the relations between the different metrics measuring the amount of perturbation and the average performance of all 3 models on all tested datasets on languages using a latin script.
    }
    \label{fig:latin_metrics}
\end{figure*}

Languages using Chinese characters and derivatives obtain a relatively weaker correlation with local perturbations.
Figure~\ref{fig:chinese_metrics} illustrates the perturbation to performance curve while only taking into account languages using Chinese characters as their script, compared to those using the Latin script in Figure~\ref{fig:latin_metrics}.

A few major divergences from the global trend are present. 
First, the average compression ratio is under 1, meaning that the tokenizer \textit{adds} to the sequence length on average.
While counter-intuitive, this is caused by the fact that the vast majority of Chinese characters' tokenization defaults to tokenizing the character directly, thus yielding almost no compression.
The tokenizer adds a few special characters for the Transformer model to use, yielding \textit{longer} sequences on average than the raw text.
This can be verified by the fact that, unlike with other scripts, subword-perturbations are sufficient to explore almost the whole spectrum of local perturbations, which would only be possible if most subwords were of length 1.

While the phrase shuffling perturbations seem to behave as expected, it seems that text written in chinese script are especially resilient to neighbour flipping.
We compare the performance of Chinese character scripts and Latin scripts in Figure~\ref{fig:latin_metrics} and find that Chinese scripts are, on average, more resilient to perturbations, going from an average score of ~$0.18$ to ~$0.08$ while the Latin Script performance drops all the way to an aggregate score of ~$0.03$.

\newpage

\end{document}